\documentclass[preprint,12pt]{elsarticle}

 \newcommand{\ra}{\rightarrow}
\newcommand{\pa}{\partial}
\newcommand{\scst}{\scriptscriptstyle}
\newcommand{\ti}{\times}
\newcommand{\D}{\displaystyle}
\newtheorem{definition}{Definition}[section]
\newtheorem{lemma}{Lemma}[section]

\newtheorem{example}[lemma]{Example}
\newtheorem{theorem}[lemma]{Theorem}
\newtheorem{algorithm}[lemma]{Algorithm}

\begin{document}

\begin{frontmatter}
\title{On the Cohomology of 3D Digital Images}
\author{Rocio Gonzalez--Diaz}
\ead{rogodi@us.es}
\ead[url]{http://www.personal.us.es/rogodi}
\author{Pedro Real}
\ead{real@us.es}
\ead[url]{http://www.pdipas.us.es/r/real}
\address{Depto. de Matem\'{a}tica Aplicada I, Escuela Superior de
Ingenier\'{\i}a Inform\'{a}tica, Universidad
de Sevilla, Avda. Reina Mercedes, s/n, 41012, Sevilla (Spain)}

\begin{abstract}
We propose a method for computing
the cohomology ring of three--dimensional (3D) digital binary--valued pictures.
We  obtain
the cohomology ring of a  3D digital binary--valued  picture $I$, via a simplicial complex $K(I)$
topologically representing (up to isomorphisms of pictures) the picture $I$.
The usefulness of a
simplicial description of the ``digital'' cohomology ring of
 3D digital binary--valued pictures is tested by means of a small program
visualizing the different steps of the method. Some examples
concerning topological thinning, the visualization of
representative (co)cycles of
   (co)homology generators and the computation of the cup
   product on the cohomology of simple pictures are showed.
\end{abstract}

\end{frontmatter}

\section{Introduction}

  The homology groups (given in terms of number of
connected components, holes and cavities in the 3D digital picture),
the digital Euler characteristic or the digital fundamental group
are well--known operations  in Digital Topology
\cite{Ros81,Kon89}. All of them can be considered as translations
into the discrete setting of classical continuous topological
invariants. In order to prove that a digital topology operation
$\pi_{\scst D}$ (associated with a continuous operation
$\pi_{\scst C}$) correctly reflects the topology of digital
pictures considered as Euclidean spaces, the main idea is to
associate  a ``continuous analog'' $C(I)$ with the digital picture
$I$. In most cases,  each digital picture  $I$ is
associated with  a polyhedron $C(I)$
\cite{Kon89,KRR92,KKM90,ADFQ00}. It is clear that $C(I)$ ``fills
the gaps'' between the black points of $I$ in a way that strongly
depends on the grid and the adjacency relations chosen for the
digital picture $I$. Recent attempts to enrich the list of
computable digital topological invariants in such a way can be
found in \cite{KI01}.

In this paper,
  starting from a 3D  digital binary--valued picture $I$,
a simplicial complex $K(I)$ associated with $I$ is constructed,
in such a way that
 an isomorphism of pictures is equivalent to a
simplicial homeomorphism of the corresponding simplicial
complexes. Therefore, we are able to define the digital cohomology
ring of $I$ with coefficients in a commutative ring $G$, as the classical cohomology ring
of $K(I)$ with coefficients in $G$ (see \cite{Mun84}).
In order to compute this last algebraic object,
it is crucial in our method to``connect'' the chain complex $C(K(I))$ canonically associated to $K(I)$ and
its homology $H(K(I))$, via an special chain equivalence \cite{Mun84}: a chain contraction
\cite{McL95}. We will obtain this goal in several steps.
Using the technique of simplicial
collapses \cite{For99}, we topologically thin $K(I)$, obtaining a
smaller simplicial complex $M_{\scst top}K(I)$ (with the same homology as $K(I)$)
and a chain contraction connecting their respective chain complexes
$C(K(I))$ and $C(M_{\scst top}K(I))$. The following step is
the construction of a  chain contraction from $C(M_{\scst top}K(I))$
to its homology $H(M_{\scst top}K(I))$.
Having all this information at hand, it is easy to compute
the digital cohomology ring of $I$ for a given commutative ring $G$.  In this way,
 cohomology rings  are computable topological invariants
which can be used
 for
``topologically'' classifying (up to isomorphisms of pictures)
and distinguishing (up to cohomology ring level) 3D
digital binary--valued pictures.

A small program for
visualizing these cohomology aspects in the case $G={\bf Z}/{\bf Z}2$
has been designed by the authors and developed by
others\footnote{ The 1st version was programmed by J.M. Berrio,
F. Leal and M.M. Maraver \cite{BGLLR01}; the 2nd version by
F.Leal. http://www.us.es/gtocoma/editcup.zip.}. This
software allows us to test in some simple examples the
potentiality and topological acuity of the method.

We deal with digital pictures derived from a
tessellation of three--space by truncated octahedra. This is equivalent to using a
body--centered--cubic--grid whose grid points are the points $(x,y,z)\in {\bf Z}^3$ in
which $x\equiv y\equiv c$ (mod $2$) (see  \cite{Kov84}).
The only Voronoi adjacency relation on this grid is $14$--adjacency.
Using this adjacency, it is straightforward to associate to a digital picture
$I$, a unique simplicial complex $K(I)$
(up to isomorphisms of pictures) with the same topological information as $I$.
This grid is important in medical imaging applications due to its
outstanding topological properties and its higher contents of
symmetries.
One advantage of the voxels in this grid is that they are more ``sphere--like'' than
the cube, so that the volumetric data represented on this grid need
fewer samples that on Cartesian cubic grid \cite{tmg01}.

Since the objects considered in this paper are
embedded in ${\bf R}^3$ then the homology groups  vanishes for dimensions greater than $3$
and they are torsion--free for dimensions $0$, $1$ and $2$
 (see \cite[ch.10]{AH35}).
 The {\em $q$th Betti number} is defined as the rank of the $q$th
homology group. In general, the $0$th Betti number is the number
of connected components, the $1$st and $2$nd Betti numbers have
intuitive interpretations as the number of independent
non--bounding loops and the number of independent non--bounding
shells. According to the Universal Coefficient Theorem for Homology,
the Betti numbers are independent of the group of
coefficients (see \cite[ch. 7]{Mun84}).
Moreover, since the homology groups are torsion--free,
the cohomology groups with coefficients in $G$ are isomorphic to
 the homology groups with coefficients also in $G$ (see \cite[ch. 5]{Mun84}).
Therefore, for simplicity we can consider that
 the ground ring is ${\bf Z}/{\bf Z}2$  throughout the paper.
 Nevertheless,
 all the procedure we explain here, is valid for any commutative ring $G$.

The paper is organized as follows. In Section 2,
the technique associating a simplicial complex to a 3D  digital binary--valued picture is detailed.
 In Section 3,
  we explain a procedure
for computing the cohomology ring of general simplicial complexes. In Section 4,  we
introduce the notion of digital cohomology ring of a 3D  digital  binary--valued  picture and
we show
some examples concerning the visualization of
representative (co)cycles of
   (co)homology generators and the computation of the cup product on the cohomology of
   simple pictures. Finally, Section 5 is devoted to conclusions and comments.

\section{From Digital Images to Simplicial Complexes}

\noindent {\bf  Digital Images}. We follow the terminology given in \cite{KRR92} for
representing
digital pictures. A {\em 3D  digital binary--valued picture space} (or, briefly, DPS) is a triple
$(V,\beta, \omega)$, where $V$ is the set of grid points in a 3D
grid and each of $\beta$ and $\omega$ is a set of closed straight
line segments joining pairs of points of $V$. The set $\beta$
(resp. the set $\omega$) determines the neighbourhood relations
between black points (resp. white points) in the grid.  A {\em 3D
digital binary--valued picture} is a quadruple $I = (V,\beta, \omega ,
B),$ where $(V,\beta, \omega)$ is a DPS and $B$ (the set of black
points) is a finite subset of $V.$

An {\em
isomorphism} of a DPS $ (V_1, \beta_1, \omega_1)$ to a DPS $(V_2,
\beta_2, \omega_2)$ is a homeomorphism $h$ of the Euclidean
$3$--space to itself such that $h$ maps $V_1$ onto
 $V_2$, each
$\beta_1$-adjacency onto a $\beta_2$-adjacency and each
$\omega_1$-adjacency onto an $\omega_2$-adjacency, and $h^{-1}$
maps each $\beta_2$-adjacency onto a $\beta_1$-adjacency and each
$\omega_2$ adjacency onto an $\omega_1$-adjacency.
An {\em isomorphism of a
picture} $I_1 = (V_1, \beta_1, \omega_1, B_1)$ to a picture $I_2
= (V_2, \beta_2, \omega_2, B_2)$ is an isomorphism of the DPS
$(V_1, \beta_1, \omega_1)$ to the DPS $(V_2, \beta_2, \omega_2)$
that maps $B_1$ onto $B_2$.

 The DPS used in this paper, is the
  3D {\em  body--centered cubic grid}  (BCC grid) \cite{KRR92}: The grid
 points ${\mathcal V}$ are the points  $(a,b,c)\in {\bf Z}^3$
  such that $a\equiv b\equiv c$ (mod $2$). The 14--neighbours of a grid point $p$ with
 coordinates $(a,b,c)$ are: $(a\pm 2,b,c)$,  $(a,b\pm 2,c)$, $(a,b,c\pm 2)$,
$(a\pm 1,b\pm 1,c\pm 1)$.

\begin{figure}[t!]
\begin{center}\includegraphics[width=8cm]{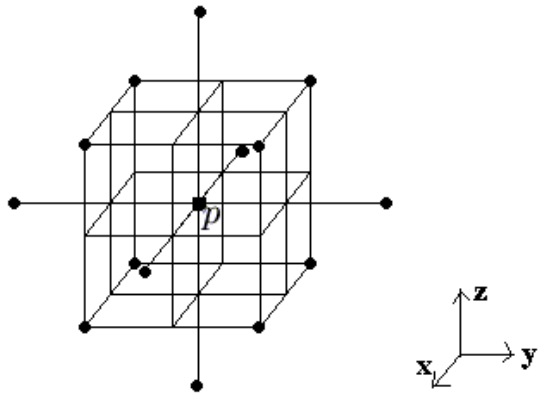}
\end{center}
 \caption{The $14$--neighbours of a grid point $p$ of the
  BCC grid.}\label{1}
\end{figure}

\noindent {\bf Simplicial Complexes}. The four types of non--empty simplices in ${\bf R}^3$ are:
  a  $0$--simplex which is a vertex, a $1$--simplex which is an edge, a
 $2$--simplex which is a triangle and  a $3$--simplex which is a
 tetrahedron. In general, considering an ordering on a vertex set $V$, a $q$--simplex with
$q+1$ affinely independent vertices $v_0<\cdots<v_q$ of $V$ is the convex hull of these points,
denoted by $\langle
v_0,\dots,v_q\rangle$.
 If $i<q$, an $i$--{\em face} of $\sigma$ is
 an $i$--simplex  whose vertices are in the set
 $\{v_0,\dots,v_q\}$. A {\em facet} of $\sigma$ is a $(q-1)$--face of it.
A simplex is {\em shared} if it is a face of more than one
simplex. Otherwise, the simplex is {\em free} if it belongs to
one higher dimensional simplex, and {\em maximal} if it does not
belong to any.

 A {\em simplicial complex} $K$ is a collection of simplices such
 that every face of a simplex of $K$ is in $K$ and the
 intersection of any two  simplices of $K$ is a face of each of them or empty.
    The set of all the $q$--simplices of $K$ is denoted by $K^{(q)}$.
       A subset $K'\subseteq K$ is a {\em subcomplex} of $K$ if it is a
 simplicial complex itself.

Let $K$ and $L$ be simplicial complexes and let  $|K|$ and $|L|$
be the subsets of ${\bf R}^{\scst d}$ that are the union of
simplices of $K$ and $L$, respectively. Let $f: K^{(0)}\ra
L^{(0)}$ be a map such that whenever the vertices $v_0,\dots,v_n$
of $K$ span a simplex of $K$, the points $f(v_0),\dots, f(v_n)$
are vertices of a simplex of $L$. Then $f$ can be extended to a
continuous map $g:|K|\ra |L|$ such that if $x=\sum t_iv_i$ then
$g(x)=\sum t_if(v_i)$. The map $g$ is called a {\em simplicial
homeomorphism} if $f$ is bijective and the points $f(v_0),\dots,
f(v_n)$ always span a simplex of $L$.

{\bf Simplicial Representations}. Given a  3D digital binary--valued
picture $I=({\mathcal V}, 14,14, B)$ on the BCC grid, there is a process to
uniquely associate  a $3$--dimensional simplicial complex
$K(I)$. This simplicial complex is constructed on the triangulation of the Euclidean $3$--space determined
by the previous $14$--neighbourhood relation.
The {\em simplicial representation} $K(I)$ of the
digital picture $I$ is described as follows:
consider the lexicographical ordering on ${\cal V}$
(if $v=(a,b,c)$ and $w=(x,y,z)$ are two points of ${\cal V}$, then $v<w$ if
$a<x$, or $a=x$ and $b<y$, or $a=x$, $b=y$ and $c<z$).
The vertices (or $0$--simplices)  of $K(I)$ are the points of $I$.
The $i$--simplices
of $K(I)$ ($i\in \{1,2,3\}$) are constituted by the different sorted
sets of $i$ $14$--neighbour black points of $I$
(analogously, we
could construct another simplicial complex whose $i$--simplices
are  the different sets of $i$ $14$--neighbour white points of
$I$).

\begin{figure}[t!]
\begin{center}\includegraphics[width=8cm]{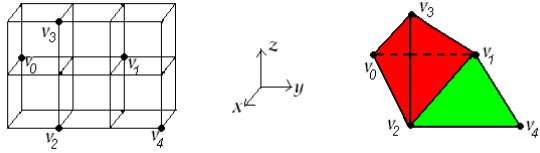}\end{center}
 \caption{On the left, the black points of the digital picture $J$ and, on the right,
  the simplicial representation $K(J)$.}\label{2}
\end{figure}

\begin{example}
Consider the digital picture $J=({\cal V},14,14,B)$ where $B$ is
the set
$\{v_0=(-1,-1,1),v_1=(-1,1,1), v_2=(0,0,0),v_3=(0,0,2),v_4=(0,2,0)\}\,;$
then $K(J)$ is the simplicial complex with set of maximal simplices
 $\{\langle v_0,$ $v_1,$ $v_2,$ $v_3\rangle,$ $\langle v_1,$ $v_2,$ $v_4\rangle\}$ (see Figure \ref{2}).
\end{example}

In the next section, we give a satisfactory algorithmic solution
 to the problem of the computation of the cohomology
ring of finite simplicial complexes. This positive solution together with the
naive simplicial construction described above
 will allow us  to
``cohomologically control'' 3D digital binary--valued pictures (up to
isomorphisms of pictures), since the following result holds.

\begin{theorem}
  Two digital binary--valued pictures,
$I_1 =$ $({\mathcal V}, 14,14,B_1)$ and $I_2=({\mathcal V}, 14,$ $14, B_2)$,
   are isomorphic
if and only if their simplicial representations $K(I_1)$ and
$K(I_2)$ are simplicially homeomorphic.
\end{theorem}

The proof of this theorem  is straightforward and left to the reader.

\section{Computing the Cohomology Ring of Simplicial Complexes}\label{basic}

First of all,  we briefly explain the main concepts from
Algebraic Topology we use in this paper.
 Our terminology follows Munkres book \cite{Mun84}.
 In the next  subsections, we
 reinterpret classical
methods in Algebraic Topology in terms of
chain contractions \cite{McL95}
 that
will enable us to design an algorithm for computing the
cohomology ring of general
 simplicial complexes.

\noindent  {\bf Chains and Homology}.
Let $K$ be a simplicial complex.
A $q$--{\em chain} $a$  is a formal sum of simplices of $K^{(q)}$.
Since the group of coefficient is ${\bf Z}/{\bf Z}2$,  a $q$--chain can be seen as a subset of
$q$--simplices of $K$; the sum of two $q$--chains $c$ and $d$ is the symmetric difference
of the two sets $c\cup d$ and $c\cap d$.
The $q$--chains form a group
 with respect to the component--wise addition mod 2; this group is the
 {\em $q$th chain group} of $K$, denoted by $C_q(K)$. There is a chain group for every
 integer $q\geq 0$, but for a complex in ${\bf R}^3$,
 only the ones for $0\leq q\leq 3$ may be non--trivial.
The {\em boundary} of a $q$--simplex $\sigma=\langle
v_0,\dots,v_q\rangle$ is the collection of all its facets which is a $(q-1)$--chain:
$\partial_q
  (\sigma)=\sum_{i=0}^q\langle v_0,\dots,\hat{v}_i,\dots, v_q\rangle$,
   where
 the hat means that $v_i$ is omitted.
 By
 linearity, the boundary operator $\partial_q$ can be extended to
 $q$--chains.
The collection of boundary operators connect the chain groups $C_q(K)$
into the {\em chain complex} $C(K)$:
$\;\cdots\stackrel{\partial_4}{\rightarrow} C_3(K)\stackrel{\partial_3}{\rightarrow}
C_2(K)\stackrel{\partial_2}{\rightarrow}
C_1(K)\stackrel{\partial_1}{\rightarrow}
C_0(K)\stackrel{\partial_0}{\rightarrow}0$.
A $q$--chain $a\in C_q(K)$ is called a
$q$--{\em cycle} if $\pa_q  (a)=0$.
  If $a=  \pa_{q+1} (a')$ for some $a'\in C_{q+1}(K)$ then $a$
 is called a $q$--{\em boundary}.
  We denote the groups of $q$--cycles
 and $q$--boundaries by $Z_q$ and $B_q$ respectively.
  An essential property of the boundary operators is that
 the boundary of every boundary is empty, $\partial_q\partial_{q+1}=0$. This implies that
 $B_q\subseteq Z_q$ for $q\geq 0$.
 Define the {\em $q$th  homology group} to be the quotient group
 $Z_q/B_q$, denoted by $H_q(K)$.
  Given  $a\in
 Z_q$,  the coset $a+B_q$ is the {\em homology  class} of $H_q(K)$ determined by $a$.
  We denote this class by  $[a]$.
For a complex $K$ in ${\bf R}^3$,
 only $H_q(K)$ for $0\leq q\leq 2$ may be non--trivial.

\noindent  {\bf Cochains and Cohomology}.
With each simplicial complex $K$, we have associated a sequence of abelian groups called
its homology groups. Now, we associate with $K$ another sequence of abelian groups called its
{\em cohomology groups}. They are geometrically much less natural than the homology groups.
Their origins lie in algebra rather than in geometry; in a certain algebraic sense, they are
``dual'' to the homology groups.

Let  $K$ be a simplicial complex. The group of $q$--{\em cochains}
of $K$ with coefficients in ${\bf Z}/{\bf Z}2$ is the group
$C^q(K)=\{c:C_q(K)\ra {\bf Z}/{\bf Z}2$  such that
$c$ is a homomorphism$\}$.
Observe that a $q$--cochain $c$ can be defined on the
$q$--simplices of $K$ and it is naturally extended to $C_q(K)$.
Therefore, a $q$--cochain can be expressed as a formal sum of
elementary cochains $\sigma^*: C_q(K)\ra {\bf Z}/{\bf Z}2$ whose value is $1$ on
the $q$--simplex $\sigma\in K$ and $0$ on all other $q$--simplices of $K$.
The boundary operator $\partial_{q+1}$ on $C_{q+1}(K)$ induces the {\em coboundary operator}
$\delta_{q}: C^q(K)\ra C^{q+1}(K)$ via $\delta_{q} (c)=c\partial_{q+1}$,
so that $\delta_q$ raises
dimension by one.
The collection of coboundary operators connect the cochain groups $C^q(K)$
into the {\em cochain complex} $C^*(K)$:
$C^0(K)
\stackrel{\delta_0}{\rightarrow}C^1(K)\stackrel{\delta_1}{\rightarrow}
C^2(K) \stackrel{\delta_2}{\rightarrow} C^3(K) \stackrel{\delta_3}{\rightarrow}\cdots.$
We define $Z^q(K)$ to be the kernel of $\delta_q$ and
$B^{q+1}(K)$ to be its image.
These groups are called the group of $q$--{\em cocycles} and $q$--{\em coboundaries}, respectively.
Noting that $\delta_q^2=0$ because $\partial_q^2=0$, define
the $q$th {\em cohomology group},
$H^q(K)=Z^q(K)/B^q(K)$  for $q\geq 0$.

The cochain complex $C^*(K)$ is an algebra with
the {\em cup product} $\smile: C^p(K)\ti C^q(K)\ra
C^{p+q}(K)$ given by:
$$(c\smile c')(\sigma)=
c (\langle
v_{0},\dots,v_{p}\rangle)\bullet c'(\langle
v_p,\dots,v_{p+q}\rangle)\,$$
where $\sigma=\langle
v_0,\dots,v_{p+q}\rangle$ is a $(p+q)$--simplex and $\bullet$ is the natural
product defined on ${\bf Z}/{\bf Z}2$ \cite[p. 292]{Mun84}.
It induces an operation
$\smile: H^p(K)\ti H^q(K)\ra H^{p+q}(K)$, via $[c]\smile [c']=[c\smile c']$,
that is bilinear,
associative, commutative (up to a sign if the ground ring is not ${\bf Z}/{\bf Z}2$),
independent of the ordering of the vertices of $K$
and topological invariant (more concretely, homotopy--type invariant) \cite[p. 289]{Mun84},
since the coboundary formula
$\delta_{p+q}(c\smile c')=\delta_p (c)\smile c'+c\smile \delta_q (c')$ holds
for any $c\in C^p(K)$ and $c'\in C^q(K)$.

\begin{example}
Consider the complex $K$ pictured in Figure \ref{3} which is obtained from a
 triangulation of a torus.

\begin{figure}[t!]
\begin{center}\includegraphics[width=8cm]{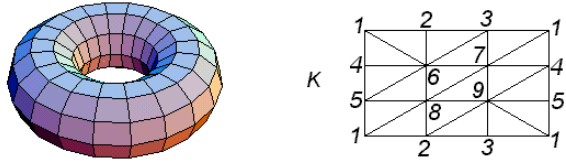}\end{center}
 \caption{A triangulation of a torus.}
\label{3}\end{figure}

It is easy to check that the two $1$--chains
$a=\langle 3,7\rangle+\langle 7,9\rangle+\langle 3,9\rangle$
and $b=\langle 3,7\rangle+\langle 6,7\rangle+\langle 6,8\rangle
+\langle 8,9\rangle+\langle 3,9\rangle$
(see Figure \ref{4})
are $1$--cycles. For example, $\partial_1(a)=\langle 3\rangle+\langle 7\rangle+\langle 7\rangle+\langle
9\rangle+\langle 3\rangle+\langle 9\rangle=0$.
Moreover, $a$ and $b$ are homologous:
$\partial_2(\langle6,7,8\rangle+\langle 7,8,9\rangle)=a+b$.
On the other hand, $c=\langle 2,3\rangle^*+
\langle 3,6\rangle^*+\langle 6,7\rangle^*+\langle 7,8\rangle^*+\langle 8,9\rangle^*
+\langle 2,9\rangle^*$ and $d=\langle 4,5\rangle^*+
\langle 5,6\rangle^*+\langle 6,8\rangle^*+\langle 7,8\rangle^*+\langle 7,9\rangle^*
+\langle 4,9\rangle^*$  are two $1$--cocycles. To check this,
we have to  verify that
$\delta_1(c)$ and $\delta_1(d)$
vanishes on all the $2$--simplices of $K$.
For example $\delta_1(c)(\langle 2,3,6\rangle)=c(\partial_2(\langle 2,3,6\rangle))=
c(\langle 2,3\rangle)+c(\langle 2,6\rangle)+c(\langle 3,6\rangle)=0$.
To check that both $1$--cocycles are not coboundaries is a more difficult task since
we have to verify that $\delta_2(f)\neq c,d$ for any $f$ being a $2$--cochain.

The cup product of $c$ and $d$ is a new $2$--cocycle $c\smile d$.
By direct computation, we have that $c\smile d=\langle 6,7,8\rangle^*$.
We obtain this by
applying it on all the $2$--simplices of $K$. For example,
$(c\smile d)(\langle 6,7,8\rangle)=c(\langle 6,7\rangle)\bullet d(\langle 7,8\rangle)=1$
and $(c\smile d)(\langle 7,8,9\rangle)=c(\langle 7,8\rangle)\bullet d(\langle 8,9\rangle)=0$.

\begin{figure}[t!]
\begin{center}\includegraphics[width=8cm]{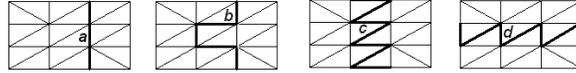}\end{center}
 \caption{On the left, the $1$--cycles $a$ and $b$, and on the right, the $1$--cocycles $c$ and $d$.}\label{4}
\end{figure}
\end{example}

The example illustrates that while  we can think of a $1$--cycle
as being a closed curve, the best way to think of a $1$--cocycle is a picket fence.

\noindent {\bf Chain Contractions}. In a more general framework,
a {\em chain complex} ${\cal C}$ is a sequence
$\cdots \stackrel{\pa_{4}}{\longrightarrow}C_3
\stackrel{\pa_{3}}{\longrightarrow}C_2\stackrel{\pa_2}{\longrightarrow}
C_{1}\stackrel{\pa_{1}}{\longrightarrow}C_0\stackrel{\pa_{0}}{\longrightarrow}0 $
of abelian groups $C_q$ and homomorphisms
$\pa_q$, indexed with the non--negative integers, such that for all $q$,
$\pa_q\pa_{q+1}=0\,.$  The {\em $q$th  homology group} is the quotient group
 Ker $\partial_q/$ Im $\partial_{q+1}$, denoted by $H_q({\cal C})$.
Let ${\cal C}=\{C_q,\pa_q\}$ and  ${\cal C'}=\{C'_q,\pa'_q\}$ be
two chain complexes. A {\em chain map} $f: {\cal C}\ra {\cal C'}$
is a family of homomorphisms $\{f_q:C_q\ra C'_q\}_{q\geq 0}$ such that
$\pa'_q
 f_q=f_{q-1}\pa_q\,.$ A
chain map $f: {\cal C}\ra {\cal C'}$ induces a homomorphism $f_*:
H({\cal C})\ra H( {\cal C'})$.

  Let us emphasize that a fundamental
notion here is that of chain contraction.
\begin{definition} \cite{McL95}
A {\em chain contraction}  of a chain complex ${\cal C}$ to
another chain complex ${\cal C'}$ is a set of three
 homomorphisms $(f,g,\phi)$
 such that:
 \begin{itemize}
 \item $f: {\cal C}\ra {\cal C'}$ and $g: {\cal C'}\ra {\cal C}$ are  chain maps.
 \item  $f g$ is the identity map of ${\cal C'}$.
 \item $\phi: {\cal C}\ra {\cal C}$ is a chain homotopy of the identity map $id_{\scst \cal C}$ of ${\cal C}$
 to $gf$, that is, $\phi  \partial+\partial \phi=id_{\scst \cal C}+g f$.
 \end{itemize}
\end{definition}
Important properties of chain contractions are:
\begin{itemize}
\item ${\cal C'}$
has fewer or the same number of generators than ${\cal C}$.
\item ${\cal C}$ and ${\cal C'}$ have isomorphic homology groups \cite[p. 73]{Mun84}.
\end{itemize}

Let us recall that the key  of our method for computing the cohomology ring of
chain complexes, is the
construction of
chain contractions $(f,g,\phi)$
of a given chain complex ${\cal C}$ to another chain complex ${\cal H}$
(isomorphic to the homology of
${\cal C}$).
In this case,  for each cycle $a\in {\cal C}$, the chain
$f(a)\in {\cal H}$
determines the homology class of $a$. Conversely, for each $\alpha\in {\cal H}$ (which corresponds to
a homology class of $H({\cal C})$),
 $g(a)\in {\cal C}$
determines a representative cycle of it. Finally, if $a\in{\cal C}$ is a boundary, then
$a'=\phi(a)$ is a chain in ${\cal C}$ such that $\partial(a')=a$.

\subsection{Topological Thinning}\label{topthinning}

 Topological thinning is an important preprocessing operation in
 Image Processing. The aim is to shrink a digital picture to a
 smaller, simpler picture which retains a lot  of the significant
 information of the original.Then, further processing or analysis
 can be performed on the shrunken picture.

There is a well--known process for thinning a simplicial complex
using simplicial collapses \cite{Bjo95}. Suppose $K$ is a
simplicial complex, $\sigma\in K$ is a maximal simplex
  and $\sigma'$ is a free facet of $\sigma$.
  Then,  $K$ {\em simplicially collapses} onto $K-\{\sigma', \sigma\}$.
  An important property of this process is that there exists an
 explicit chain contraction of  $C(K)$ to
$C(K-\{\sigma', \sigma\})$ \cite{For99}.
 More generally, a {\em simplicial collapse} is any sequence of
 such operations.
  A {\em thinned} simplicial complex
  $M_{\scst top}K$ is a subcomplex of $K$ with the condition
that
  all the faces
 of the maximal simplices of $M_{\scst top}K$ are shared.
  Then, it is
 obvious that it is no longer possible to collapse.

   The following algorithm computes $M_{\scst top}K$ (first step) and a chain contraction
 $(f_{\scst top},$ $g_{\scst top},$ $\phi_{\scst top})$
 of $C(K)$ to $C(M_{\scst top}K)$ (second step). In particular,
 recall that this means that the (co)homology of $K$ and $M_{\scst top}K$ are
 isomorphic. Each step of the algorithm runs in time at most $O(m^2)$ if $K$ has $m$
simplices.

\begin{algorithm}Topological Thinning Algorithm.

First step: Simplicial collapses.
  \begin{tabbing}
  {\sc Input:} {\tt A simplicial complex $K$}.\\
     {\tt Initially,} $M_{\scst top}K:=K$, {\em collapse} $:=(\;)$,
     {\em pair} $:=$ {\em True}.\\
    {\bf  While} \= {\em pair} {\tt is} {\em True} {\tt do}\\
    \> {\em pair} $:=$ {\em False}.\\
    \> {\bf For} \= {\tt each $\sigma\in M_{\scst top}K$ do}\\
    \> \>{\bf If} \= {\tt $\sigma$ is maximal with a free facet $\sigma'$ in
    $M_{\scst top}K$ then}\\
    \>\> \>$M_{\scst top}K:=K-\{\sigma',\sigma\}$,\\
    \>\> \>{\em collapse} $:=$  $ (\sigma',\sigma)\,\cup$ {\em collapse},\\
   \>\>\> {\em pair} $:=$ {\em True}.\\
   \>\> {\bf End if}.\\
   \> {\bf End for}.\\
   {\bf End while}.\\
   {\sc Output:} \= {\tt the simplicial complex: } $M_{\scst top}K$\\
    \> {\tt and the sorted set of
   simplices:} {\em collapse}.
\end{tabbing}

Second step: the computation of the chain contraction.
  \begin{tabbing}
  {\sc Input:} \= {\tt The simplicial complexes $K$ and $M_{\scst top}K$}\\
  \> {\tt and the sorted set }
  {\em collapse} $=(\sigma_1',\sigma_1,\dots,\sigma'_n,\sigma_n)$.\\
      {\tt Initially,} \= {\tt  $f_{\scst top}(\sigma):=\sigma$,
      $\phi_{\scst top}(\sigma):=0$ for each $\sigma\in K$;}\\
      \> {\tt  and
$g_{\scst top}(\sigma):=\sigma$ for each $\sigma\in M_{\scst top}K$.}\\
    {\bf For} \= $i=1$ {\tt to} $i=n$ {\tt do}\\
\> $f_{\scst top}(\sigma'_i):=f_{\scst top}(\partial\sigma_i+\sigma'_i)$,\\
\> $\phi_{\scst top}(\sigma'_i):=\sigma_i+\phi_{\scst top}(\partial\sigma_i+\sigma'_i)$,\\
\>  $f_{\scst top}(\sigma_i):=0$.\\
{\bf End for}.\\
   {\sc Output:} {\tt the chain contraction
   $(f_{\scst top}, g_{\scst top},\phi_{\scst top})$ of $C(K)$ to
   $C(M_{\scst top}K)$}.
\end{tabbing}
\end{algorithm}

\begin{figure}[t!]
\begin{center}\includegraphics[width=8cm]{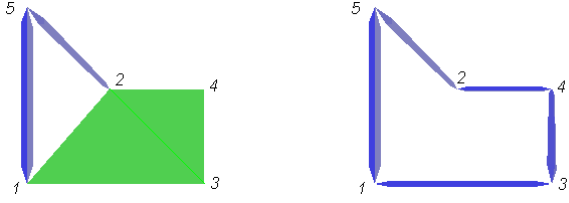}\end{center}
 \caption{The simplicial complexes $L$ (on the left) and
 $M_{\scst top}L$ (on the right).}\label{5}
\end{figure}

\begin{example}
Consider the simplicial complex $L$ whose set of maximal simplices is
$\{\langle 1,5\rangle,\langle 2,5\rangle,\langle 1,2,3\rangle, \langle 2,3,4\rangle\}$
(see Figure \ref{5}).
Applying the first part of the algorithm above we  have that
$M_{\scst top}L$ $= \{\langle 1,3\rangle,$ $ \langle 3,4\rangle,$ $\langle 2,4\rangle,$
$\langle 1,5\rangle,$ $\langle 2,5\rangle\}$
and
$\mbox{{\em collapse} } =( \langle 2,3\rangle, \langle 2,3,4\rangle,
\langle 1,2\rangle,$ $\langle 1,2,3 \rangle).$
The stages of the second part of the algorithm is showed in the following table:

\begin{tabular}{c|c|c}
\hline
$\sigma $&$ f_{\scst top}(\sigma) $& $\phi_{\scst top}(\sigma)$\\
\hline
$\langle 2,3 \rangle$&$ f_{\scst top}\langle 2,4 \rangle+ f_{\scst top}\langle 3,4 \rangle$&$
\langle 2,3,4 \rangle+\phi_{\scst top}\langle 2,4 \rangle+ \phi_{\scst top}\langle 3,4 \rangle$\\
&$=
\langle 2,4 \rangle+ \langle 3,4 \rangle$&$
=\langle 2,3,4 \rangle$\\
\hline
$\langle 2,3,4 \rangle$ &$0$&$0$\\
\hline
$\langle 1,2 \rangle$&$f_{\scst top}\langle 1,3 \rangle+f_{\scst top}\langle 2,3 \rangle
$
&$\langle 1,2,3\rangle+ \phi_{\scst top}\langle 1,3 \rangle+\phi_{\scst top}\langle 2,3 \rangle$\\
&$=\langle 1,3 \rangle+\langle 2,4 \rangle+\langle 3,4 \rangle$
&$=\langle 1,2,3\rangle+ \langle 2,3,4 \rangle$\\
\hline
$\langle 1,2,3 \rangle$ &$0$&$0$\\
\hline

\end{tabular}
\end{example}

\subsection{``Algebraic Thinning''}

Having obtained the thinned complex $M_{\scst
top}K$, we next construct a
chain contraction $(f_{\scst alg},g_{\scst alg},
\phi_{\scst alg})$ of the chain complex
$C(M_{\scst top}K)$ to its homology. This step can be considered
as a
 thinning at algebraic level (for this reason we call it
 ``algebraic thinning'').
We compute $(f_{\scst alg},g_{\scst
alg},\phi_{\scst alg})$
 interpreting  the
``incremental algorithm'' \cite{DE95} for computing homology
groups in ${\bf R}^3$,  in terms of chain contractions. As we will see later, the design of
an algorithm for computing the cohomology ring of $K$, will be possible thanks to the information
saved in the chain contraction  of $C(K)$ to its homology constructed before.

Let $(\sigma_1,\dots,\sigma_m)$ be a sorted set of all the
simplices of  $K$ with the property that
any subset $\{\sigma_1,\dots,\sigma_i\}$, $i\leq m$, is a
subcomplex of it. Algorithm \ref{alghom}
computes a chain complex ${\cal
H}$ with a set of generators $h$, and a chain contraction
$(f_{\scst alg},g_{\scst alg}, \phi_{\scst alg})$ of $C(K)$ to ${\cal H}$. Initially, $h$ is empty.
 In the $i$th step  of the algorithm, the simplex $\sigma_i$ is added to the
 subcomplex $\{\sigma_1,\dots,\sigma_{i-1}\}$ and then, a homology class is created or
 destroyed.
 If $f_{\scst alg}\partial(\sigma_i)=0$ then $\sigma_i$ ``creates'' a homology class.
 Otherwise, $\sigma_i$
 ``destroys'' one homology class ``involved'' in the expression of
  $f_{\scst alg}\partial(\sigma_i)$.
 At the end of the algorithm,
 ${\cal H}$ is a chain complex isomorphic to the homology of $K$.

\newpage

\begin{algorithm}\label{alghom} Algebraic Thinning Algorithm

\begin{tabbing}
{\sc Input:} {\tt The  sorted set} $(\sigma_1,\dots,\sigma_m)$.\\
{\tt Initially,} \=  {\tt  $f_{\scst alg}(\sigma):=0$,
      $\phi_{\scst alg}(\sigma):=0$ for each $\sigma\in K$; and $h:=\{\;\}$.}\\
{\bf For } \= {\tt  $i=1$ to $i=m$  do}\\
\> {\bf If } \=  {\tt $ f_{\scst alg}\partial(\sigma_i)= 0$ then}\\
\>\> $h:=h\cup \{\sigma_i\}$\\
\>\> $ f_{\scst alg}(\sigma_i):=\sigma_i$.\\
 \> {\bf Else } \=  {\tt take any one $\sigma_j$
 of $f_{\scst alg}\partial(\sigma_i)$, then}\\
\>\>  $ h:=h-\{\sigma_j\}$,  \\
\>\> {\bf For } \= {\tt  $k=1$ to $k=m$} {\tt  do}\\
\>\>\> {\bf If} \= {\tt $\sigma_j$ appears in the expression of $f_{\scst alg}(\sigma_k)$ then}\\
\>\>\>\> $f_{\scst alg}(\sigma_k):= f_{\scst alg}(\sigma_k)+ f_{\scst alg}\partial(\sigma_i)$\\
\>\>\>\> $\phi_{\scst alg}(\sigma_k):= \phi_{\scst alg}(\sigma_k)+\sigma_i+ \phi_{\scst alg}\partial(\sigma_i)$\\
\> \>\>{\bf End if}.\\
\>\>{\bf End for}.\\
\>{\bf End if}.\\
{\bf End for}.\\
 {\bf For } \= {\tt each $\sigma\in h$ do} \\
 \> $g_{\scst alg}(\sigma):=\sigma+\phi_{\scst alg}\partial(\sigma)$.\\
{\bf End for }.\\
{\sc Output:} {\tt the chain contraction $(f_{\scst alg}, g_{\scst alg}, \phi_{\scst alg})$ of
$C(K)$ to ${\cal H}$.}
\end{tabbing}
\end{algorithm}

The output of the algorithm  allows us to determine both a
 representative cycle
 for each homology class and the homology class for each
 cycle.
 Moreover, for any
$q$--boundary $a$ on $K$ we can obtain a $(q+1)$--chain
$a'=\phi_{\scst alg}(a)$ on $K$ such that $a=\partial
(a')$.

Concerning to the complexity, suppose
$K$ has $m$
simplices. In the $i$th step of the algorithm ($1\leq i\leq m$), we have to evaluate
$\partial \sigma_i$. The
number of simplices involved in $\partial \sigma_i$ is fewer or the same
 than the dimension of  $\sigma_i$ which is at most $3$. On the other hand,
 the number of elements involved in the formulae for
$f_{\scst alg}\partial \sigma_i$ and $\phi_{\scst alg}\partial \sigma_i$ is $O(3
m)=O(m)$. If $\partial \sigma_i\neq 0$, we have to update
$f_{\scst alg}\partial \sigma_k$ and $\phi_{\scst alg}\partial \sigma_k$ for $1\leq k\leq m$, so the total cost of these
operations is $O(m^2)$.
 Therefore, the total algorithm runs in time at most $O(m^3)$.

Let us observe that composing  the chain contraction $(f_{\scst
top},g_{\scst top}, \phi_{\scst top})$ of
$C(K)$ to $C(M_{\scst top}K)$, described in the
previous subsection, with  $(f_{\scst alg},g_{\scst
alg},\phi_{\scst alg})$ of $C(M_{\scst top}K)$
 to ${\cal H}$ (isomorphic to $H(K)$), we get a
new chain contraction
$(f_{\scst alg}
f_{\scst top},$ $g_{\scst top} g_{\scst alg},$
$\phi_{\scst top}+g_{\scst top} \phi_{\scst
alg} f_{\scst top})$  of $C(K)$ to ${\cal H}$.

\begin{figure}[t!]
\begin{center}\includegraphics[width=8cm]{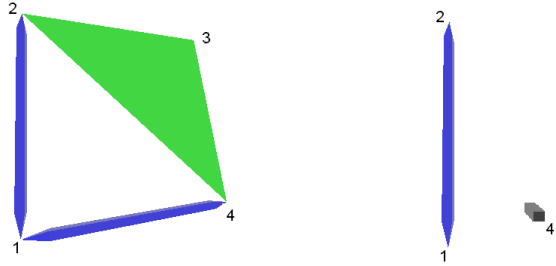}\end{center}
 \caption{The   complex $L$ (on the left) and the generators of
 ${\cal H}_{\scst L}$ (on the right).}\label{6}
\end{figure}

\begin{example}
Let $L$ be the simplicial complex  showed in  Figure \ref{6}.
The intermediate stages of the algorithm are:

$$\begin{array}{c|c|c|c|c|c|c|c|c|c|c|c}
\hline
i&  0& 1 & 2 & 3 & 4 & 5 & 6 & 7 & 8 & 9 & 10\\
\hline
\sigma && \langle 1 \rangle & \langle 2 \rangle & \langle 3 \rangle & \langle 2,3 \rangle &
\langle 4 \rangle & \langle 3,4 \rangle & \langle 1,4 \rangle & \langle 1,2 \rangle
& \langle 2,4 \rangle & \langle 2,3,4 \rangle\\
\hline
f_{\scst alg}(\langle 1 \rangle)&0&\langle 1 \rangle&&&&&&\langle 4 \rangle&&&\\
\phi_{\scst alg}(\langle 1 \rangle)&0&&&&&&&\langle 1,4 \rangle&&&\\
f_{\scst alg}(\langle 2 \rangle)&0&&\langle 2 \rangle&&\langle 3 \rangle&&\langle 4 \rangle&&&&\\
\phi_{\scst alg}(\langle 2 \rangle)&0&&&&\langle 2,3 \rangle&&\langle 2,3 \rangle+\langle 3,4 \rangle&&&&\\
f_{\scst alg}(\langle 3 \rangle)&0&&&\langle 3 \rangle&&&\langle 4 \rangle&&&&\\
\phi_{\scst alg}(\langle 3 \rangle)&0&&&&&&\langle 3,4 \rangle&&&&\\
f_{\scst alg}(\langle 2,3 \rangle)&0&&&&&&&&&&\\
\phi_{\scst alg}(\langle 2,3 \rangle)&0&&&&&&&&&&\\
f_{\scst alg}(\langle 4 \rangle)&0&&&&&\langle 4 \rangle&&&&&\\
\phi_{\scst alg}(\langle 4 \rangle)&0&&&&&&&&&&\\
f_{\scst alg}(\langle 3,4 \rangle)&0&&&&&&&&&&\\
\phi_{\scst alg}(\langle 3,4 \rangle)&0&&&&&&&&&&\\
f_{\scst alg}(\langle 1,4 \rangle)&0&&&&&&&&&&\\
\phi_{\scst alg}(\langle 1,4 \rangle)&0&&&&&&&&&&\\
f_{\scst alg}(\langle 1,2 \rangle)&0&&&&&&&&\langle 1,2 \rangle&&\\
\phi_{\scst alg}(\langle 1,2 \rangle)&0&&&&&&&&&&\\
f_{\scst alg}(\langle 2,4 \rangle)&0&&&&&&&&&\langle 2,4 \rangle&0\\
\phi_{\scst alg}(\langle 2,4 \rangle)&0&&&&&&&&&&\langle 2,3,4 \rangle\\
f_{\scst alg}(\langle 2,3,4 \rangle)&0&&&&&&&&&&\\
\phi_{\scst alg}(\langle 2,3,4 \rangle)&0&&&&&&&&&&\\
\hline
\end{array}$$

Finally, $h=\{\langle 4\rangle,\langle 1,2\rangle\}$,
$g_{\scst alg}(\langle 4 \rangle):= \langle 4 \rangle$  and
$g_{\scst alg}(\langle 1,2 \rangle):= \langle 1,2 \rangle+\langle 1,4 \rangle+\langle 2,3 \rangle+
\langle 2,4 \rangle\,.$

Summing up, the  output  is a chain contraction
 $(f_{\scst alg},g_{\scst
alg},\phi_{\scst alg})$ of $C(L)$ to the chain complex ${\cal H}_{\scst L}$
(isomorphic to $H(L)$)
with set of generators $h=\{\langle 4\rangle, \langle 1,2\rangle\}$.
In particular, we obtain that
$H_0(L)\simeq {\bf Z}/{\bf Z}2$,
  $H_1(L)\simeq {\bf Z}/{\bf Z}2$
  and $H_2(L)=0$.
\end{example}

\subsection{Computing the Cohomology  Ring}

After applying in order topological and  algebraic thinning to the
simplicial complex $K$, we are able to compute the multiplication table on the
cohomology.
Since the ground ring is a field, the homology and cohomology groups of $K$
are always isomorphic.

Let $(f,g,\phi)$ be a chain contraction of $C(K)$ to ${\cal H}$
(where ${\cal H}$ is a chain complex isomorphic to $H(K)$ and to $H^*(K)$),
and let $h=\{\alpha_1,\dots,\alpha_p\}$ be
 a set of generators of ${\cal H}$
obtained using the algorithms
explained before. Then
$\alpha^*_i f:C_q(K)\ra {\bf Z}/{\bf Z}2$ (where
$\alpha^*_i(\alpha_j)=1$ if $j=i$ and $0$
 otherwise)
is a representative cocycle of the cohomology class corresponding to
$\alpha_i$, for $1\leq i\leq p$.

Let $\alpha$ and $\beta$ be two elements of $h$, then the cup product
of the cohomology classes corresponding to
$\alpha$ and $\beta$ can be computed as follows:

\begin{algorithm}  The cup product of two classes of cohomology.
\begin{tabbing}
{\sc Input:} \= {\tt the elements $\alpha$ and $\beta$}\\
\> {\tt and the set of generators $h=\{\alpha_1,\dots,\alpha_p\}$ of ${\cal H}$.}\\
{\tt Initially,} $\lambda_k:=0$ {\tt for} $1\leq k\leq p$ {\tt and}
$cup:=0$.\\
{\bf For }\= {\tt  $k=1$ to $k=p$} {\tt  do}\\
\> $\D \lambda_k:= (\alpha^*f\smile \beta^*f)g(\alpha_k)$.\\
{\bf End for.}\\
$cup:=\sum_{k=1}^{p}\lambda_k\alpha_k$.\\
{\sc Output:} {\tt the chain} $cup$.
 \end{tabbing}
 \end{algorithm}
Observe that the complexity of this algorithm for computing $\alpha\smile \beta$ is $O(m^4)$.
Moreover, if we are interested in computing the cohomology ring of $K$,
we have to
apply the algorithm above to all the  pairs $(\alpha_i,\alpha_j)$, $1\leq i\leq j\leq p$ (since the
cup product is commutative, $\alpha_i\smile \alpha_j=\alpha_j\smile \alpha_i$).
Then, the  algorithm for computing the cohomology ring of $K$
will run in time at most $O(m^6)$ if $K$ has $m$ simplices.

Let us note that the cohomology
ring of $K$ is
     not suitable in general  for topological classification tasks. This is  due to the
     fact that determining whether two rings are isomorphic or not by means of their
     respective multiplication
     tables is an extremely difficult computational question. In
     order to avoid this problem, we will put the information of the cup product table into
     a different form.

If we restrict our interest in  simplicial complexes  are embedded in ${\bf R}^3$, observe that
the possible non--trivial
cup products are the ones $\alpha\smile\beta$ where both $\alpha$ and $\beta$ are
elements of $h$  corresponding to  cohomology classes of dimension $1$.

In order to design a new algorithm for computing the cup product in a way  that we
can determine whether two cohomology rings are isomorphic or not by means of their
     respective multiplication
     tables, we need to define a new concept.
Given a  chain contraction $(f,g,\phi)$ of $C(K)$
  to a chain complex ${\cal H}$ (isomorphic
to $H^*(K)$) with set of generators $h$,
and a simplex $\sigma=\langle v_0,v_1,v_2\rangle $ of dimension $2$, suppose that
$\{\alpha_1,\dots,\alpha_p\}$ is the set of elements of $h$ of dimension $1$,
$f(\langle v_0,v_1\rangle)=\sum_{i\in I} \alpha_i$ and $f(\langle v_1,v_2\rangle)=\sum_{j\in J} \alpha_j$ where
$I$ and $J$ are subsets of the set $\{1,2,\dots,p\}$. Define
$(f\odot f)(\sigma)=\sum_{i\in I}\sum_{j\in J}(\alpha_i,\alpha_j)$. This definition can
be extended to $2$--chains by linearity.

\begin{algorithm}
\label{cupalg} Cup Product Algorithm
\begin{tabbing}
{\sc Input:} \= {\tt A simplicial complex $K$}\\
\> {\tt and a chain contraction $(f,g,\phi)$ of $C(K)$}\\
\> {\tt  to a chain complex ${\cal H}$ (isomorphic
to $H^*(K)$)}\\
\> {\tt  with set of generators $h$.}\\
{\tt Initially,} \=
{\tt $q:=$ the number of elements of $h$ of dimension $2$,}\\
\> $b_i:=0$ {\tt for $1\leq i\leq q$ and $M:=(\,)$.}\\
{\bf For }\= {\tt  $i=1$ to $i=q$} {\tt  do}\\
\> $b_i:= (f\odot f)g(\alpha_i)$.\\
{\bf End for.}\\
$M:=(b_1,\dots,b_q)$.\\
{\sc Output:} {\tt The sorted set} $M$.
 \end{tabbing}
 \end{algorithm}

Let $\{\alpha_1,\dots,\alpha_p\}$ be the set of elements of $h$ of dimension $1$ and
 $\{\beta_1,\dots,\beta_q\}$ the ones of dimension $2$.
Each
$b_i$, $1\leq i\leq q$, is of the form
$ \sum\lambda^i_{jk}(\alpha_j,\alpha_k)$
where the sum is taken over the set $\{(j,k): 1\leq j,k\leq p\}$ and
 $\lambda^i_{j,k}=(\alpha_j^*f\smile\alpha_k^*f)g(\beta_i)$
could be $0$ or $1$. Since the cup product is commutative, we have that
$\lambda^i_{jk}=\lambda^i_{kj}$.
Therefore, the output of this algorithm can be put into a matrix form ${\cal M}$ of
(cohomology classes $\beta_i$, $1\leq i\leq q$) $\times$
(pairs  of cohomology clases $(\alpha_j,\alpha_k)$, $1\leq j\leq k\leq p$).
The column of ${\cal M}$ corresponding to the pair
$(\alpha_j,\alpha_k)$, $1\leq j\leq k\leq p$,
 gives the value of the cup product $\alpha_j\smile \alpha_k$.
This  algorithm for
computing the matrix ${\cal M}$  runs in time at most
$O(m^4)$ if $K$ has $m$ simplices.

     From the diagonalization $D$ of the matrix ${\cal M}$,
      a first cohomology invariant $HB_1(K)$ for distinguishing
non--homeomorphic simplicial complexes with isomorphic
(co)homology groups appears. We define this cohomology number in order to have a handy numerical
tool for distinguishing 3D digital pictures.
\begin{definition}
Given a simplicial complex $K$, the integer $HB_1(K)$ is a cohomology invariant
defined as the rank of the matrix ${\cal M}$.
\end{definition}

\section{A First Approach to the Digital Cohomology Ring}\label{section4}

Since an isomorphism of pictures on the 3D body centered cubic grid is equivalent to a
simplicial homeomorphism of the corresponding simplicial
representations, we are able to define the digital cohomology ring of
$I$ with coefficients in ${\bf Z}/{\bf Z}2$ as the cohomology ring of $K(I)$
with coefficients in ${\bf Z}/{\bf Z}2$. Moreover, the following definitions hold:

\begin{definition}
Given a  digital picture $I= ({\mathcal V}, 14,14, B)$, the digital
cohomology ring of $I$ with coefficients in ${\bf Z}/{\bf Z}2$ is defined as  the
cohomology ring of $K(I)$ with coefficients in ${\bf Z}/{\bf Z}2$. The cohomology invariant
$HB_1(I)$ is defined as $HB_1(K(I))$.
\end{definition}

In the previous sections, we have showed that it is possible to compute the digital
cohomology ring of $I$ with coefficients in ${\bf Z}/{\bf Z}2$.
The steps of the method are: first, we construct the simplicial complex $K(I)$. Second,
we topologically thin $K(I)$, obtaining a
smaller simplicial complex $M_{\scst top}K(I)$ and a chain contraction
$(f_{\scst top},g_{\scst top},\phi_{\scst top})$ of $C(K(I))$
to  $C(M_{\scst top}K(I))$.
Third, we compute ${\cal H}$ which is isomorphic to $H(I)$ and a chain contraction
$(f_{\scst alg},g_{\scst alg},\phi_{\scst alg})$ of
$C(M_{\scst top}K(I))$ to ${\cal H}$.
Four,
 we calculate the cohomology ring of $I$ via the cohomology ring of $M_{\scst top}K(I)$
 and the invariant $HB_1(I)$ via $HB_1(M_{\scst top}K(I))$,
 using the chain contractions constructed before.
All the information obtained
 in this way is useful for
topologically classifying
and distinguishing  binary
3D digital pictures.

\subsection{Some Examples}\label{section5}

In order to show examples of the
computation and visualization of the cohomology ring of simple
pictures,  we expose a small prototype called
EditCup. We use a free program for building 3D worlds. In our
case, a world is a particular 3D  simplicial complex $K$
representing a digital picture $I$ considering the
$14$--adjacency. A way for distinguishing the different maximal
simplices of a simplicial
representation is by using different colours: red for tetrahedra,
green for triangles, blue for edges, and black for vertices.

All the computations are done considering ${\bf Z}/{\bf Z}2$ as the ground ring.
For visualizing (co)chains, the simplices on which a given (co)chain
is non--null, are lighted in a different
 color.
On the other hand, the ``visualization'' of any
(co)homology class on $K$ is given
 by lighting  the simplices  of $K$
on which the representative cochain of this class is non--null.
Moreover, the ``visualization'' of any
(co)homology class on the original  3D digital binary--valued picture
$I$ could be  given
 by lighting  the points of $I$ such that the corresponding vertices span simplices
on which the representative cochain of this class is non--null.

\begin{figure}[t!]
\begin{center}\includegraphics[width=8cm]{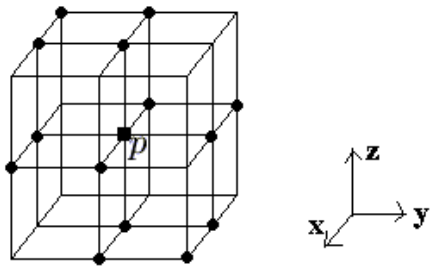}\end{center}
 \caption{The $14$--neighbours of a grid point $p$ of the (14,14)--DPS.}\label{7}
\end{figure}

The DPS used in these examples, that we call $(14,14)$--DPS, is
 $({\bf Z}^3, 14,14)$, in which the underlying grid is the set of
points with integer coordinates in the Euclidean $3$--space $E^3$
and the $14$--neighbours of a grid point (black or white) with
integer coordinates $(x,y,z)$ are: $(x\pm 1,y,z)$,
$(x,y\pm 1,z)$, $(x,y,z\pm 1)$, $(x+1,y-1,z)$, $(x-1,y+1,z)$, $(x+1,y,z-1)$, $(x-1,y,z+1)$,
$(x,y+1,z-1)$, $(x,y-1,z+1)$, $(x+1,y+1,z-1)$, $ (x-1,y-1,z+1)$ (see Figure \ref{7}).
The $(14,14)$--DPS and the BCC
 grid are isomorphic DPSs: each grid point $(x,y,z)$ of the $(14,14)$--DPS can be associated to
a point $(a,b,c)$ via the formula: $(a,b,c)=(x+y-2z,-x+y,-x-y))$.

\begin{figure}[t!]
\begin{center}\includegraphics[width=8cm]{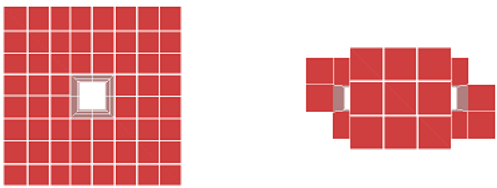}\end{center}
\caption{The
volumetric representation of the pictures $I$ (on the left) and $J$ (on the right).}\label{8}
\end{figure}

 Let us consider now the following pictures: a
  torus  $I=({\bf Z}^3,14,14,B_{\scst I})$ and a
  wedge of two topological circles and a topological $2$--sphere
   $J=({\bf Z}^3,14,14,B_{\scst J})$  (see Figure \ref{8}).
In the volumetric representation of the picture
$I$ (resp. $J$), we use voxels with centres   the points $B_{\scst I}$ (resp. $B_{\scst J}$).
  It is clear that the
(co)homology groups of $I$ are
  isomorphic to those of $J$. They are ${\bf Z}/{\bf Z}2$, ${\bf Z}/{\bf Z}2\oplus {\bf Z}/{\bf Z}2$ and
  ${\bf Z}/{\bf Z}2$ of dimension $0$, $1$ and $2$, respectively. So,
  the (co)homology information is not enough for topological distinguishing both pictures.

\begin{figure}[t!]
\begin{center}\includegraphics[width=8cm]{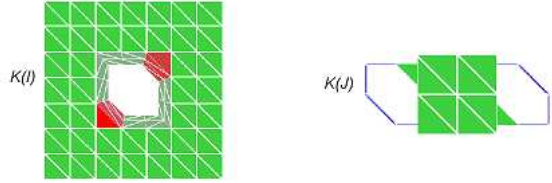}\end{center} \caption{The
simplicial complexes $K(I)$ (on the left) and  $K(J)$ (on the right).}\label{9}
\end{figure}

Let $K(I)$ and $K(J)$ be the simplicial representations of $I$ and $J$ respectively (see Figure \ref{9}).
In order to compare the cohomology ring of both pictures,
the first step is the computation of chain contractions
$(f_{\scst I},g_{\scst I},\phi_{\scst I})$ of $C(K(I))$ to ${\cal H_{\scst I}}$ and
$(f_{\scst J},g_{\scst J},\phi_{\scst J})$ of $C(K(J))$ to ${\cal H_{\scst J}}$ using the
topological and algebraic thinning algorithms explained before, where ${\cal H_{\scst I}}$
(resp. ${\cal H_{\scst J}}$) is a chain complex isomorphic to the (co)homology of
$I$ (resp. $J$).

\begin{figure}[t!]
\begin{center}\includegraphics[width=8cm]{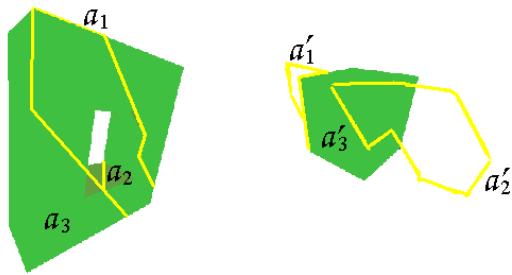}\end{center}\caption{The
cycles $a_1$, $a_2$ and $a'_1$, $a'_2$ (in yellow);  and $a_3$
and $a'_3$ (in green).}\label{10}
\end{figure}

\begin{figure}[t!]
\begin{center}\includegraphics[width=8cm]{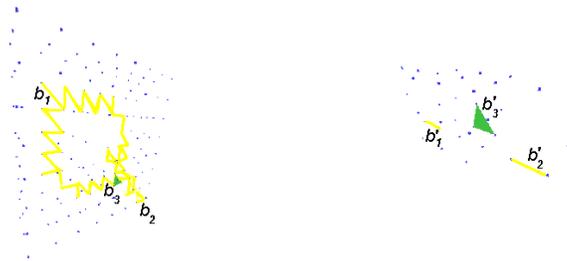}\end{center}
 \caption{The  cocycles  $b_1$,  $b_2$ and  $b'_1$, $b'_2$ (in yellow); and $b_3$ and
 $b'_3$ (in green).}\label{11}
\end{figure}

Let us denote by $\alpha_1$ and $\alpha_2$ (resp.  $\alpha'_1$ and $\alpha'_2$)
the generators of ${\cal H_{\scst I}}$ (resp. ${\cal H_{\scst J}}$) of dimension $1$
and $\alpha_3$ (resp. $\alpha'_3$) the generator of ${\cal H_{\scst I}}$
(resp. ${\cal H_{\scst J}}$) of dimension $2$.
Let us also denote by $a_i$ the representative cycles
of the generators of $H(I)$ (that is, $a_i=g_{\scst I}(\alpha_i)$); and by
$a'_i$ the same of $H(J)$.
  We visualize these  cycles on $K(I)$ and $K(J)$
   in Figure \ref{10}. In Figure \ref{11},  the  representative cocycles $b_i$ (resp.
   $b'_i$) obtained via the formula $b_i=\alpha_i^*f_{\scst I}$ (resp.
   $b'_i=\alpha_i'^*f_{\scst J}$)
     of the generators of  $H^*(I)$ (resp. $H^*(J)$) are shown. Recall that we do it
     by lighting  the simplices
on which the  cochains is non--null.

The output of Algorithm \ref{cupalg}
for ${\cal H}_{\scst I}$ and ${\cal H}_{\scst J}$ are
$M_{\scst I}=((\alpha_1,\alpha_2)+(\alpha_2,\alpha_1))$ and
$M_{\scst J}=(0)$, respectively.
The matrices corresponding to
the cohomology rings of the pictures  $I$ and $J$ are:
\begin{center}
\begin{tabular}{c|ccc}
$I$&$(\alpha_1,\alpha_1)$&$(\alpha_1,\alpha_2)$&$(\alpha_2,\alpha_2)$\\
\hline $\alpha_3$&$0$&$1$&$0$
\end{tabular}\hspace{1cm}
\begin{tabular}{c|ccc}
$J$&$(\alpha'_1,\alpha'_1)$&$(\alpha'_1,\alpha'_2)$&$(\alpha'_2,\alpha'_2)$\\
\hline $\alpha'_3$&$0$&$0$&$0$
\end{tabular}
\end{center}
Therefore, $HB_1(I)=1$ and $HB_1(J)=0$.
We conclude that $K(I)$ and $K(J)$ are not homeomorphic (more precisely, we conclude that they are not
homotopy equivalent), then $I$ and $J$ are not isomorphic.

\begin{figure}[t!]
\begin{center}\includegraphics[width=8cm]{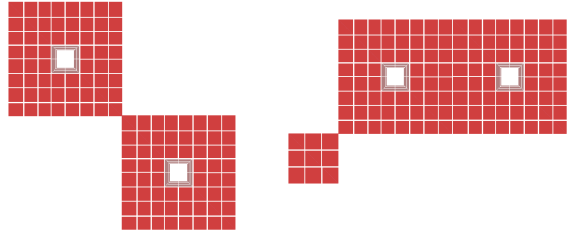}\end{center}
 \caption{The
pictures $A$ (on the left) and $B$ (on the right).}\label{12}
\end{figure}

Let us expose another example (see Figure \ref{12}): the picture $A$
is a wedge of two torus; the picture $B$ consists in a wedge
of a sphere and a genus--$2$ torus (a sphere with two handles and two holes).
Both pictures have $1$ connected component, $4$ holes and
$2$ cavities.

\begin{figure}[t!]
\begin{center}\includegraphics[width=8cm]{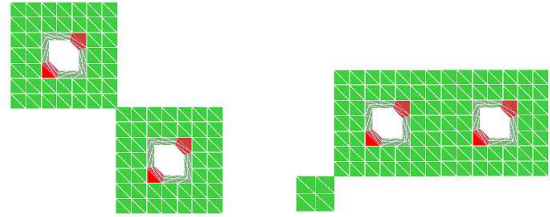}\end{center} \caption{The
simplicial complexes $K(A)$ (on the left) and $K(B)$ (on the right).}\label{13}
\end{figure}

\begin{figure}[t!]
\begin{center}\includegraphics[width=8cm]{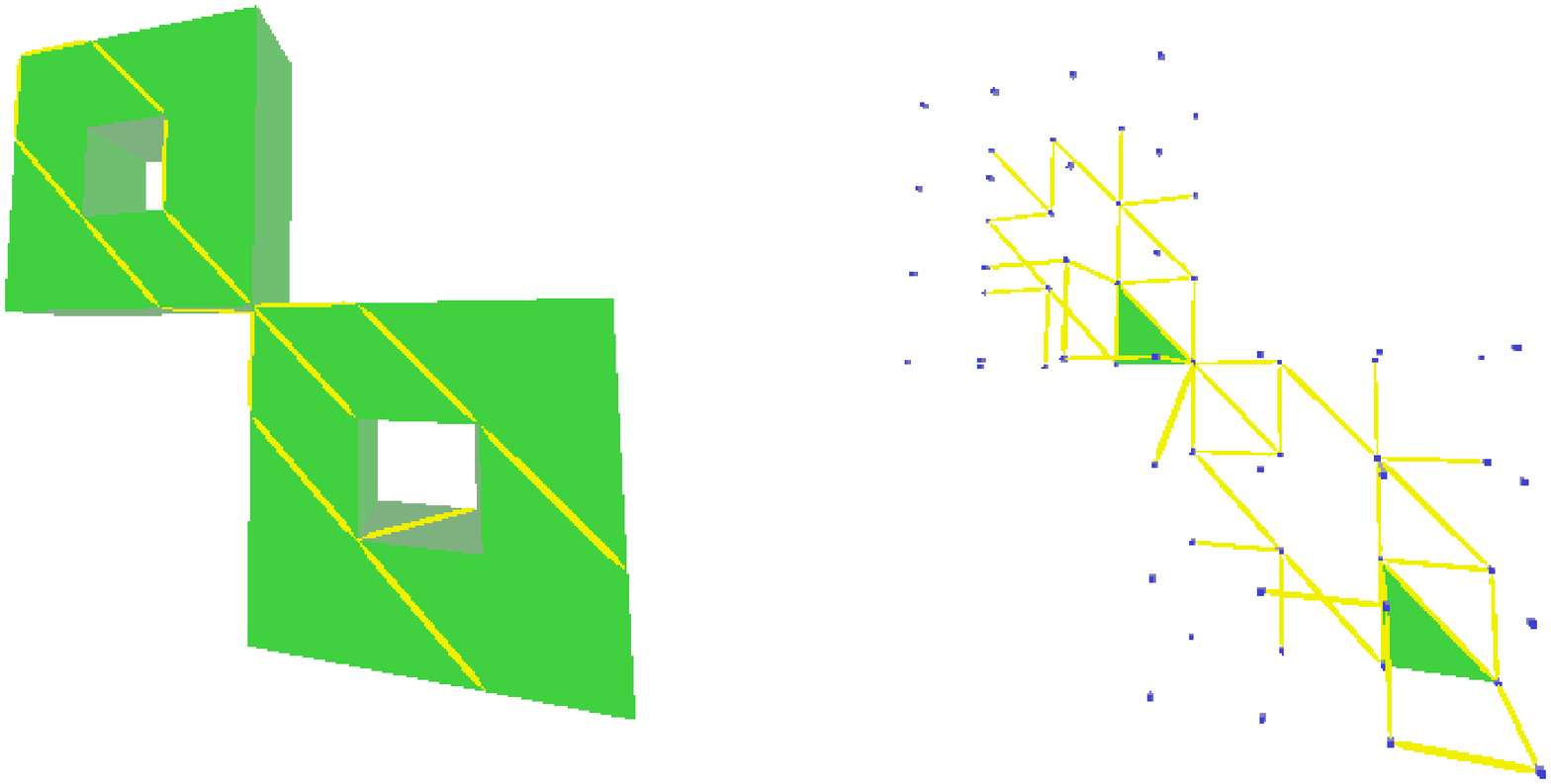}\end{center} \caption{On the left
(resp. on the right),
the representative
cycles (resp. cocycles) of the generators of $H(A)$ (resp. of $H^*(A)$).}\label{14}
\end{figure}

\begin{figure}[t!]
\begin{center}\includegraphics[width=8cm]{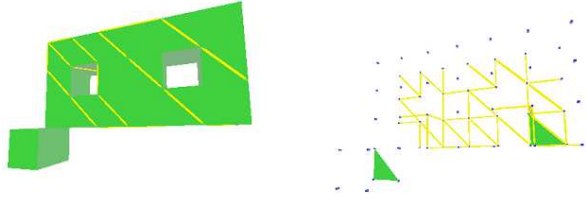}\end{center}
 \caption{On the left (resp. on the right),
the representative
cycles (resp. cocycles) of the generators of $H(B)$ (resp. of $H^*(B)$).}\label{15}
\end{figure}

The simplicial representations of $A$ and $B$, $K(A)$ and $K(B)$ are showed in Figure \ref{13}.
In Figure \ref{14} (resp. Figure \ref{15}), the representative
cycles and cocycles of the generators of
the (co)homology of $A$ (resp.  $B$) are showed.

Let us denote by $\alpha_i$ (resp. $\alpha'_i$), $i=1,2,3,4$, the
generators of ${\cal H}_{\scst A}$ (resp. ${\cal H}_{\scst B}$) of dimension 1;
and by $\beta_i$
(resp. $\beta'_i$), $i=1,2$, the
generators of ${\cal H}_{\scst A}$ (resp. ${\cal H}_{\scst B}$) of dimension 2.
All of them are obtained using the algebraic thinning
algorithm explained in the previous section.
The output of Algorithm \ref{cupalg} for ${\cal H}_{\scst A}$ and ${\cal H}_{\scst B}$
is $M_{\scst A}=((\alpha_1, \alpha_2)+(\alpha_1, \alpha_3)+
(\alpha_2, \alpha_1)+(\alpha_3, \alpha_1), (\alpha_3,\alpha_4)+(\alpha_4,\alpha_3))$ and
$M_{\scst B}=(0, (\alpha_1\,\alpha_4)+(\alpha_2,\alpha_3)
+(\alpha_3,\alpha_2)+(\alpha_4,\alpha_1))\,.$
Therefore, the matrices corresponding to the
cohomology rings of  $A$ and $B$ are:
\begin{center}
\begin{tabular}{c|cccccccccc}
$A$&$(1, 1)$&$(1, 2)$
&$(1, 3)$
&$(1, 4)$
&$(2, 2)$
&$(2, 3)$
&$(2, 4)$
&$(3, 3)$
&$(3, 4)$
&$(4, 4)$
\\
\hline $\beta_1$&$ 0   $&$ 1 $&$ 1 $&$ 0 $&$ 0 $&$ 0 $&$ 0$&$ 0$&$ 0$&$ 0$      \\
$\beta_2$&$ 0$&$0$&$0$&$0$&$0$&$0$&$0$&$0$&$1$&$0$
\end{tabular}\end{center}
\begin{center}
\begin{tabular}{c|cccccccccc}
$B$&$(1, 1)$&$(1, 2)$
&$(1, 3)$
&$(1, 4)$
&$(2, 2)$
&$(2, 3)$
&$(2, 4)$
&$(3, 3)$
&$(3, 4)$
&$(4, 4)$
\\
\hline $\beta'_1$    &$ 0   $&$ 0 $&$ 0 $&$ 0 $&$ 0 $&$ 0 $&$ 0$&$ 0$&$ 0$&$ 0$     \\
$\beta'_2$&$0$&$0$&$0$&$1$&$0$&$1$&$0$&$0$&$0$&$0$
\end{tabular}
\end{center}
where $(i,j)$ represents the pair $(\alpha_i,\alpha_j)$
(resp. $(\alpha'_i,\alpha'_j)$).

We conclude that $HB_1(A)=2$ and $HB_1(B)=1$, and then
$K(A)$ and $K(B)$ are not homeomorphic (more precisely, we conclude that they are not
homotopy equivalent), therefore $A$ and $B$ are not isomorphic.

\section{Conclusions and Future Work}

We have seen that there is a true algorithm for computing the digital
cohomology ring (with coefficients in ${\bf Z}/{\bf Z}2$)
of
a  3D binary picture on the BCC grid. It is also possible to compute the digital
cohomology ring of $I$ with coefficients in any commutative ring $G$,
thanks to the fact the simplicial complex $K(I)$
is embedded in ${\bf R}^3$ and, consequently, it
have torsion--free homology.
We deal here
with ${\bf Z}/{\bf Z}2$ coefficients, in order to simplify and avoiding signs in
the explanation of our algorithmic formulation,
to give an
easy geometric interpretation of "digital" cohomology classes and to work
with binary arithmetic. Moreover, there is no problem to define the
cohomology ring of $I$ with coefficients in a commutative ring $G$ as
the cohomology ring of $K(I)$ with coefficients in $G$; and the
cohomology invariant
$HB_1(I;G)$ with coefficients in $G$ as $HB_1(K(I);G)$. On the other hand, since
$HB_1(K(I);G)$ can be obtained
from
the first homology
group of  the reduced bar construction $\bar{B}(C^*(K(I);G))$ \cite{McL95}
associated to the cochain complex
$C^*(K(I);G)$ with coefficients in $G$, we will confine
ourselves to say that the rest of homology groups of this last algebraic object give rise to
more complicated cohomology invariants for a  digital binary--valued picture.

   In this paper, we talk about topological and algebraic thinning. Concerning the first one, we
do not use here well--known direct (without passing to simplicial framework)
topological thinnings of digital binary--valued pictures because we are interested in constructing
chain contractions which allow us to obtain cohomology results. Concerning the second one,
the idea of computing a chain contraction of a chain complex to its
homology has also been used in \cite{GR01} for
computing primary and secondary cohomology operations.

 Another important question is to try to improve the complexity of the algorithm computing
digital cohomology ring on the BCC grid detailed in
this paper. We do not take advantage here neither of the particular simplicial structure of the
simplicial complex $K(I)$ (determined by the BCC grid)  associated to $I$,
nor of representing in a
compressed form (without loss of information) the  3D digital picture (for example, in an octree
format). To obtain positive results in these directions and to eliminate from our algorithm the
intermediary simplicial objects will allow us to specify a more refined algorithm computing
digital cohomology on the BCC grid.

   Finally, another important question that it is necessary to deal with in a near future is to
try to generalize this work to other natural digital picture spaces.

\end{document}